%% file: MAIN.tex
\documentclass[sigconf]{acmart}

\usepackage{tcolorbox}

\usepackage[utf8]{inputenc} 
\usepackage[T1]{fontenc}    
\usepackage{hyperref}       
\usepackage{url}
\usepackage{multirow}       
\usepackage{booktabs}       
\usepackage{amsfonts}       
\usepackage{nicefrac}       
\usepackage{tabularx}
\usepackage[hide]{notes-alt} 
\usepackage{longtable}
\usepackage{supertabular}
\usepackage{makecell}
\usepackage{microtype}      
\usepackage{xcolor}         
\usepackage{tabularx}
\usepackage{graphicx}
\graphicspath{ {imgs/} }
\usepackage{svg}
\usepackage{enumitem}
\usepackage{subcaption}
\usepackage{amsthm, amsmath,algorithm2e}

\usepackage{colortbl}

\definecolor{blue}{RGB}{17,220,247}
\definecolor{purple}{RGB}{163,115,250}
\definecolor{caribbeangreen}{rgb}{0.0, 0.8, 0.6}

\definecolor{GREEN}{RGB}{84,130,53}
\newcommand{\colorit}{\cellcolor{green!15}}
\newcommand{\coloritt}{\cellcolor{orange!15}}
\newcommand{\colorg}{\cellcolor{gray!15}}

\usepackage{wrapfig}
\newcommand{\mpara}[1]{\medskip\noindent{\bf #1}}

\DeclareMathOperator*{\argmax}{arg\,max}

\newcommand{\icat}{\textsc{icat}}
\newcommand{\knn}{\textsc{knn}}
\newcommand{\mmr}{\textsc{mmr}}
\newcommand{\transfer}{\textsc{transfer}}
\newcommand{\target}{\textsc{target}}
\newcommand{\flanxl}{\textsc{flan-t5-xl}}
\newcommand{\flan}{\textsc{flan-t5-large}}
\newcommand{\ftd}{\textsc{ftd}}
\newcommand{\COT}{\textsc{Few-shot-cot}}
\newcommand{\self}{\textsc{self-ask}}
\newcommand{\autocot}{\textsc{auto-cot}}
\newcommand{\fshot}{\textsc{few-shot}}
\newcommand{\zfshot}{\textsc{zero-shot-cot}}

\newcommand{\aqua}{\textsf{AquaRat}}
\newcommand{\tab}{\textsf{TabMwp}}
\newcommand{\fin}{\textsf{FinQA}}
\newcommand{\arith}{\textsf{MultiArith}}
\newcommand{\sv}{\textsf{SvAmp}}
\newcommand{\mqa}{\textsf{MusiqueQA}}
\newcommand{\sub}{\textsf{SubQA}}
\newcommand{\strat}{\textsf{StrategyQA}}
\newcommand{\wqa}{\textsf{WQA}}

\copyrightyear{2023}
\acmYear{2023}
\setcopyright{rightsretained}

\title{In-Context Ability Transfer for Question Decomposition in Complex QA}

\begin{document}

\author{Venktesh V}
\email{v.viswanathan-1@tudelft.nl}
\affiliation{%
  \institution{Delft University of Technology}
  \country{Netherlands}
}

\author{Sourangshu Bhattacharya}
\email{ssourangshu@cse.iitkgp.ac.in}
\affiliation{%
  \institution{IIT Kharagpur}
  \country{India}
  }

  \author{Avishek Anand}
  \email{Avishek.Anand@tudelft.nl}
  \affiliation{%
    \institution{Delft University of Technology}
    \country{Netherlands}
  }

\begin{abstract}

\input{abstract-AA}

\end{abstract}
\maketitle

\input{01-introduction-AA}

\input{02-related-AA}

\input{03-framework}

\input{05-experiments-avi}

\input{06-results-AA}
\balance
\bibliographystyle{ACM-Reference-Format}
\bibliography{reference} 

\end{document}

%% file: abstract-AA.tex
Answering complex questions is a challenging task that requires question decomposition and multistep reasoning for arriving at the solution. 
While existing supervised and unsupervised approaches are specialized to a certain task and involve training, recently proposed prompt-based approaches offer generalizable solutions to tackle a wide variety of complex question-answering (QA) tasks.
However, existing prompt-based approaches that are effective for complex QA tasks involve expensive hand annotations from experts in the form of rationales and are not generalizable to newer complex QA scenarios and tasks.
We propose, \icat{} (In-Context Ability Transfer) which induces reasoning capabilities in LLMs without any LLM fine-tuning or manual annotation of in-context samples. We transfer the ability to decompose complex questions to simpler questions or generate step-by-step rationales to LLMs, by careful selection from available data sources of related tasks. 
We also propose an automated uncertainty-aware exemplar selection approach for selecting examples from \transfer{} data sources.
Finally, we conduct large-scale experiments on a variety of complex QA tasks involving numerical reasoning, compositional complex QA, and heterogeneous complex QA which require decomposed reasoning. 
We show that ICAT convincingly outperforms existing prompt-based solutions without involving any model training, showcasing the benefits of re-using existing abilities.

%% file: 01-introduction-AA.tex
\section{Introduction}

\label{sec:intro}
Answering complex questions is an active research topic that has widespread applications in finance, medical diagnosis, technical support, and more, enabling in-depth information access and advanced problem-solving~\cite{roy2022question:book}.
Complexity could arise from multiple factors, such as the need to parse heterogeneous sources \cite{chen2022finqa} or multiple documents \cite{tang-etal-2021-multi} to answer the question. 
It could also arise from the nature of the questions, such as compositional questions \cite{2wikimultihopqa, trivedi-etal-2022-musique}, which require reasoning through question decomposition to arrive at the final answer \cite{decomposing_complex_multi_hop} or numerical reasoning \cite{chen2022finqa,aqua_rat,roy2016solving}. 
An example of a compositional natural language question is \textit{``Who is the spouse of the person who voices Jarvis in Iron Man?"}. To answer this question, it must be first decomposed into relevant sub-questions, such as \textit{``Who voices Jarvis in Iron Man ?"} and \textit{``Who is the spouse of Paul Bettany?"}. To arrive at the final answer, a system must answer the sub-questions and aggregate their answers. The decomposition leads to interpretable reasoning apart from improved performance on the answering task. In numerical reasoning-based complex questions, stepwise decomposition of the reasoning process, usually called rationales, helps induce arithmetic abilities to arrive at the correct solution \cite{aqua_rat,roy2016solving}. 
Alternate approaches also involve generating arithmetic programs from the given natural language problem statement. \cite{chiang-chen-2019-semantically,amini-etal-2019-mathqa}. 

Existing methods for compositional complex QA adopt supervised \cite{decomposing_1, zhou-etal-2022-learning-decompose, min-etal-2019-multi,decomposing_complex_multi_hop, khot2021textmodular, deng2022interpretable} or unsupervised approaches \cite{perez-etal-2020-unsupervised} to decompose a complex question into simpler questions and provide the final answer by aggregating answers to the sub-questions. 
Similarly, numerical reasoning questions require supervised stepwise rationales along with the solution for fine-tuning generative models \cite{aqua_rat,cobbe2021training, geva-etal-2020-injecting}. 
More recently, a new paradigm for solving complex questions using general purpose Large Language Models (LLMs) has emerged using in-context learning~\cite{gpt3incontext} and chain-of-thought reasoning~\cite{complex_cot,wei2023chainofthought}.
Instead of fine-tuning task-specific models, these approaches focus on choosing carefully constructed natural language prompts with demonstrations and rationales.
A distinct advantage of prompting vs fine-tuning is  that the LLM is a general-purpose machine that can be used to solve multiple complex QA tasks.
In this paper, our focus is to propose a general framework that can be employed to solve multiple complex QA tasks by selecting and re-using relevant abilities from pre-existing datasets for automatic prompt construction.

\mpara{The limitations of LLM-based approaches.} 
It is inherently hard for the LLMs to perform multi-step reasoning using few-shot in-context learning with just the  \texttt{(input, output)} pairs due to limited context and lack of demonstrable  execution strategies~\cite{rae2022scaling}.
An improvement to few-shot in-context learning, therefore, is \textit{chain-of-thought} (COT) prompting that proposes to add hand-crafted natural language rationales to the prompts in addition to input and output \texttt{(input, rationale, output)}~\cite{wei2023chainofthought}.
The process of generating stepwise natural language rationales can be seen as decomposing the original problem to simpler problems that can be solved with less effort\cite{wei2023chainofthought, brown2020gpt3}.

However, there are two inherent limitations of COT for complex QA. 
Firstly, COT involves the manual effort of collecting a large set of rationales for each complex QA task. 
Existing approaches require careful hand-annotated sub-questions or natural language rationales. This limits the information contained in exemplars, as it is cumbersome to hand-annotate a large pool of samples in the training set. Additionally, different tasks require the annotator to craft different reasoning chains, resulting in non-trivial effort. 
Secondly, and more subtly, is that rationales for complex QAs are typically not in the form of question decompositions leading to lack of faithfulness and hallucinations~\cite{turpin2023language,ye2022unreliability}. 
Consequently, there have been methods that avoid hallucinations by adding hand-crafted follow-up questions derived from the original complex questions to the prompt like self-ask~\cite{self_ask}.
However, this leads to even more manual effort and is not scalable to new complex QA tasks, and different distributions. 
In the example below, using Few-shot-COT results in inaccurate results due to lack of decomposition information in its prompt (hidden due to space reasons).
\input{intro-example-mouth}

\mpara{\icat{} (In-Context Ability Transfer).} 
In this work, we propose a new approach for inducing stepwise decomposition abilities in LLMs for complex QA without expensive fine-tuning or manual annotation processes. 
Our approach involves reusing rationales in existing QA datasets, which we refer to as \transfer{} datasets.
We hypothesize that existing specialized QA datasets with rationales or decompositions might already contain instances that can be re-used as demonstrations for question decompositions.
This avoids the necessity of manual annotation of newer complex QA scenarios and tasks.
Given a new question, we propose both static and dynamic selection approaches to automatically select relevant abilities necessary for decomposing the complex question.
For dynamic selection, we provide a uncertainty aware distance metric called the frechet-term distance (\ftd{}) that has nice theoretical properties and is more robust than otherwise sensitive pointwise estimates of distance.
We conduct extensive experiments on a large number of different complex QA datasets and show that ICAT convincingly outperforms all prompting-based competitors, and in some cases is even competitive with SOTA models trained just for the task.

\mpara{Contributions.} The core contributions of our work are:
\begin{description}
    \item 
    1) We propose a new approach ICAT that re-uses rationales from similar ability datasets for effective chain-of-thought reasoning.

    \item 
    2) We propose a novel uncertainty-aware exemplar selection approach for dynamic prompt construction that can achieve competitive performance with no manual intervention.
    
    \item 
    3) We conduct extensive experiments with different exemplar selection methods to showcase the generalized performance of ICAT across multiple complex QA tasks.
\end{description}

%% file: intro-example-mouth.tex
\begin{tcolorbox}[title= Example: \COT{} vs \icat{}]
\small
\textbf{Question}: \texttt{What is the mouth of the river which serves as the mouth of the Bumping River?}

\paragraph{\textbf{\COT{}}}: [Answer]: \textcolor{red}{There is no river named Bumping River.} 

\paragraph{\textbf{ICAT}}

[Question 0]: \texttt{What is the river that serves as the mouth of the Bumping River?
}

[Answer 0]: \textsf{The Naches River}

[Question 1]: \texttt{What is the mouth of the Naches River?}

[Answer 1]: \textsf{Yakima River}

[Final Answer]: \textcolor{teal}{\textsf{Yakima River}}

\end{tcolorbox}

%% file: 02-related-AA.tex
\section{Related Work}
\label{sec:rel-work}

We divide our related work into challenges in a) question decomposition in complex question answering, b) complex QA using LLMs, and c) automatic prompt construction in LLMs.  

\subsection{Complex Question Answering and Question Decomposition}

Complex question answering is a sub-field in question answering (QA) where the notion of complexity is varied~\cite{roy2022question:book}.
The source of complexity in complex QA tasks can be due to a complex  question structure~\cite{trivedi-etal-2022-musique,aqua_rat}, the need to perform comparisons or aggregations~\cite{2wikimultihopqa,tabmwp}, mathematical operations~\cite{roy2016solving}, or having to extract from complex objects like tables and text~\cite{chen2022finqa}.
Question understanding strategies used for simple questions are therefore not applicable to complex QA tasks.
Supervised and unsupervised~\cite{perez-etal-2020-unsupervised}  approaches for complex QA often rely on a multi-hop strategy where the question representation is iteratively refined over multiple reasoning steps until arriving at the final answer~\cite{perez-etal-2020-unsupervised,multi_hop_survey}.
These approaches typically employ end-to-end training strategies where the internal process is represented in latent vector spaces and share vector representations of the questions and answers between various stages.
Alternately, question decomposition approaches employ natural language parsing to decompose a complex question into multiple sub-questions, followed by reasoning and aggregation steps \cite{decomposing_complex_multi_hop,decomposing_1, zhou-etal-2022-learning-decompose, min-etal-2019-multi,decomposing_complex_multi_hop, khot2021textmodular, deng2022interpretable}.
A more recent trend has been to use question decomposition techniques for LLMs by forgoing training models altogether~\cite{patel-etal-2022-question} and using in-context learning (ICL) using prompt engineering~\cite{icl_bayesian,gpt3incontext}. Apart from text based QA tasks, recent works like \cite{ye2023large} have explored the possibility of employing LLMs for evidence and question decomposition for table based reasoning.
In this work, we intend to use the reasoning capability of LLMs to decompose complex questions in simpler and understandable natural language sub-questions.

\subsection{Chain of Thought for Complex Question Answering}

When using LLMs to solve complex tasks like \textit{numerical, multi-step reasoning}, it has been shown that constructing prompts with stepwise execution strategy in natural language or \textit{chain-of-thought} for arriving at the final solution helps to arrive at a more accurate solution~\cite{wei2023chainofthought}.
We refer to these reasons in chain-of-thought~(COT) reasoning as \textit{rationales}.
Prior work has successfully demonstrated the applicability of COT in numerous complex reasoning tasks~\cite{aqua_rat,cobbe2021training} against vanilla ICL methods~\cite{few_shot,shin-etal-2020-autoprompt,rae2022scaling}.
However, a bottleneck of COT approaches is that the rationale has to be manually constructed.

Secondly, some studies have indicated that such explanations are unfaithful \cite{turpin2023language} to the model's reasoning process and the model ignores the COT explanation at times when arriving at the final answer. 
Alternate approaches like successive prompting \cite{dua-etal-2022-successive} propose to generate a sub-question followed by another call to answer the question. 
This leads to a computationally expensive inference phase. To arrive at a concise answer in a stepwise manner, self-ask \cite{self_ask} proposes decomposing compositional questions to simpler follow-up questions but requires hand-crafted demonstration samples. Further, inspired by retrieval augmented LLMs\cite{khandelwal2020generalization,lewis2021retrievalaugmented}, self-ask combines neural LMs with the search for arriving at more accurate answers for sub-questions. 

However, a major drawback in proposed approaches is the need for manual annotation of rationales or sub-questions. This is cumbersome and also limits the generality of the approach, as the chosen exemplar pool is limited to annotated samples. In this paper, we focus on transferring the ability to generate rationales for numerical reasoning and the ability to decompose complex questions. 
We accomplish this by relying on \transfer{} datasets with expert annotated rationales or sub-questions.

\subsection{Exemplar  Selection for ICL}

Prompt engineering is still an ongoing effort as it is a complex multi-objective optimization problem to select a few empirical observations for achieving stable performance on a wide range of test samples.
Selecting relevant examples or exemplar selection is key to improved ICL performance.
However, exemplar selection has been shown to be sensitive to presentation order, dataset, task, and models \cite{zhao2021calibrate,su2022selective, lu-etal-2022-fantastically}.  Few auto exemplar selection methods like similarity based \cite{rubin-etal-2022-learning} approaches, complexity based \cite{complex_cot} and MMR \cite{ye-etal-2023-complementary} based approaches for diverse example selection have been proposed to achieve stable performance across a wide range of tasks. However, for COT dynamic retrieval of exemplars for each test sample requires annotations of rationales/reasoning paths for the whole training set which is cumbersome. Though Auto-COT \cite{zhang2023automatic} proposes to mitigate manual annotation, it relies on zero-shot-cot for designing exemplars, which is erroneous and leads to error propagation on downstream questions. Hence, in our work, we study existing automatic or dynamic selection methods in a novel setting where we select exemplars from transfer datasets with expert rationale annotations. We also propose a new exemplar selection method and evaluate its effectiveness across a wide range of tasks.

%% file: 03-framework.tex
\section{Proposed Ability Transfer Framework}
\label{sec:framework}

In this section, we describe the proposed paradigm \icat{} for ability transfer in LLMs. 

\subsection{Complex Question Decomposition}

Given a set of complex questions with corresponding answers $D = \{(q_1,a_1),...,(q_n,a_n)\}$, where $n$ is the total size, $q_i \in Q$ (set of questions), $a \in A$ (set of answers); the question decomposition models usually attempt to generate a decomposition.
We define a decomposition for a question-answer(QA) pair $(q,a)$ as
$de(q,a) = \{(sq_1,sa_1),..., (sq_k,sa_k)\}$, $k=k_{(q,a)}$ is the number of sub-questions for a given question $(q,a)$ and $(sq_i,sa_i)$ are the pairs of sub-question and corresponding answers. The set $de = de(q,a)$ (for simplicity, we drop QA pair from the decomposition $de$ when it is obvious) can be seen as an ordered set, building up a reasoning chain to arrive at a correct answer $a \in A$. In the reasoning chain, each sub-question $sq_j$ follows the answer to the previous question $sq_i$. The final answer $a$ can be seen as an aggregation of all the answers $sa_i,...,sa_k$ to the sub-questions \cite{dua-etal-2022-successive,decomposing_complex_multi_hop}.

Supervised complex QA approaches assume annotated contexts, such as passages $C$ and a corresponding decomposition, \\ $de=\{(sq_1,sa_1)...(sq_k,sa_k)\}$ as input. 
A decomposition model $\hat{f}$ is trained on the given question and golden decomposition pairs. In multi-hop complex QA\cite{roy2022question:book}, the trained model $\hat{f}$ approximates a function $f: Q
 \times C^n \xrightarrow[]{} A\times DE$  that generates the decomposition $de\in DE$ and the answer $a\in A$ to the question $q$:
 \[f(q_i,C) = (\hat{a_i} \in A, \hat{de}_i=\{(sq_1,sa_1),..., (sq_k,sa_k)\} )\]
 where, $sa_i \subset C$. Generally, the answer $\hat{a_i}$ is an aggregation of the answers to the questions in the decomposition $\hat{de_i}$.

 Unsupervised methods do not assume access to annotated decomposition, $de$, for each question and generate a pseudo decomposition based on context $C$ or external knowledge.
 A similar formulation applies to complex multi-step numerical reasoning questions, where $sq_i$ corresponds to a step of arithmetic reasoning and $sa_i$ the corresponding answer. The final answer is obtained by aggregating all answers $sa_i,...,sa_k$ to intermediate arithmetic reasoning steps.
 In Section \ref{solution}, we elaborate on our solution framework leveraging Large Language Models. We do not rely on contexts $C$ or annotated decomposition for the questions in the evaluation datasets. We also do not incur the cost of fine-tuning language models.  We rely on the parametric memory of LLMs and induce reasoning in LLMs through few shot demonstration examples.



\input{table-datasets}

\subsection{Solution Framework}
\label{solution}
Given a complex QA dataset as \target{}, $D_t = \left\{\left(q_1^t,a_1^t\right),...,\left(q_n^t,a_n^t\right) \right\}$ (question-answer pairs), the goal is to identify a \transfer{} dataset $D_s =\ \left \{(q_1^s,a_1^s),...,(q_n^s,a_n^s)\ ; z_s\right\}$ which contain a desired ability $z_s$ required to solve the task of the \target{} dataset $D_t$.
\sloppy
Our framework accomplishes this through an automated construction of a prompt using demonstration samples  
from $D_s$. The constructed prompt is input to a $\mathbb{LLM}$  to leverage In-Context Learning (ICL) to arrive at the correct solution.
The key idea here is that, given a target question $q_i^t$ the prompt constructed from a subset of the demonstration set $D_s$, should retain the ability to induce a correct decomposition of the question: $\hat{de_i}^t = de(q_i^t)$. Hence, we formally define ability as:
 \begin{definition}{(Ability)}
        It is the characteristic $z_s$ demonstrated in a dataset, $D_s$ in the form of a reasoning chain, $de = \left\{ (sq_1^s,sa_1^s), \dots, (sq_k^s,sa_k^s)  \right\}$ required to solve a task $D_t$.
  \end{definition}
For instance, $z \ = de$ could represent the stepwise arithmetic reasoning rationales, decomposed sub-questions or the stepwise parsing of tables.  Also, in our work, the ability is a characteristic of the dataset and not any neural model.
It is important to note that the desired ability is absent in the \target{} dataset $D_t$ and the \icat{} aims to transfer this ability from another dataset $D_s$ which acts as a \textit{\transfer{} dataset}. 
Another important aspect of our method is \textit{transferability}, defined as the ability $z_s$ is said to be transferable if the exemplars (demonstration samples) $\left\{(q_1^s,a_1^s,de_1^s)..\right\} $in $D_s$  help $LLM$ generate an execution strategy/decomposition $\hat{de_i^t}$ and final answer $\hat{a_i}$ resulting in competitive performance on $D_t$.

    

In \icat{}, the ability to decompose is transferred by first selecting a set of \textbf{exemplars}: 
$\psi_S=\left\{\left(q_i^s,a_i^s,de_i^s\right)\right\}_{i=1}^m \subset D_s$ where  $ m << n$.  The exemplars are salient examples, from $D_s$ that help transfer the ability for a given target task. We discuss two exemplar selection methods in sections \ref{sec:staticselection} and \ref{sec:dynamicselection}.

\begin{tcolorbox}[title=\icat{} Prompt]
\small
Instructions: You are a [\dots]
 \\Examples: Q: $\{q_i^s\}$?  A:$\{a_i^s\}$ Decomposition: [Question1]: $sq_1^s$,[Answer1]: $sa_1^s$  [\dots] \\
 Test Q: ${q_i^t}$, Decomposition: [INS], A: [INS]
\end{tcolorbox}

In the second step, the exemplars are fed as a prompt as shown above to the $\mathbb{LLM}$ for conditional answer generation. Formally,
\[\hat{de_i}^t (\tau)  = \mathbb{LLM}\left(. | \psi_S,q_{i}^t,\right) \ ; \hat{a_i} (\tau+1) = \mathbb{LLM}\left(. | \psi_S,q_{i}^t,\hat{de_i^t}\right)  \]
where $\tau$ refers to the time step in autoregressive decoding of output from LLM.
For each question $q_i^t$, the LLM  produces a decomposition $\hat{de_i^t}$ and final answer $\hat{a_i}$ in an autoregressive fashion by aggregating answers to the sub-questions  $(sq_i^t \dots)$ in the decomposition, in  \textbf{one inference pass} of the LLM with aid of instructions in the prompt.
Both the decomposition and the final aggregated answer is obtained using the parametric memory of LLM without relying on context or annotations.

\subsection{Static Exemplar Selection}
\label{sec:staticselection}

As described above, the purpose of exemplar selection is to identify the most important exemplars for a particular task. Broadly, there can be two types of exemplar selection techniques: (1) \textbf{Static exemplar selection} - where one set of exemplars are selected for the entire task, and (2) \textbf{Dynamic exemplar selection} - where one set of exemplars are selected for each question (described in section \ref{sec:dynamicselection}).
For \icat{} static selection, we carefully select exemplars such that they cover diverse question types, such as \textit{compositional} and \textit{comparison} type questions.  For numerical reasoning-based questions, we ensure each selected exemplar is diverse and the selected set overall covers different question types such as those requiring compositional arithmetic reasoning, etc. Conceptually, the exemplars are chosen according to a utility function $f(q_i^t,e^s_i)$,
\begin{equation}
\label{eq:utility}
    u^s_i = \argmax_{i \in D_s \setminus \psi_S}\left({f\left(q_i^t,\left(\psi_S \sqcup e_i^s\right) \right)} - f\left(q_i^t, \psi_S\right)\right)
\end{equation}
Each exemplar $u^s_i = \{q_i,a_i,de_i\}$ is added to the set only if it improves the utility.
For static exemplar selection, we manually select the exemplar using coverage as our utility metric ($f$).
 To summarise the overall method, we consider coverage of question types and diversity as our utility and add an exemplar to set $\psi_S$ following Equation \ref{eq:utility}. Since the exemplars in the transfer dataset are already annotated with rationales, it requires less effort when compared to manual \COT{}.
 
The \transfer{} datasets and \target{} datasets and related abilities considered in this work are listed in Table
\ref{tab:datasets_overview}. Our goal is to identify \transfer{} datasets close to the \target{} dataset in terms of task complexity to transfer the decomposition ability from \transfer{} datasets. Hence, we compute \textit{Jensen-Shannon (JS) divergence} between unigram (token) distributions of samples in transfer and target datasets, as shown in Table
\ref{tab:datasets_overview}. We observe low divergence values except for \strat{} $\xrightarrow[]{} $ \wqa{}. We observe this because \wqa{} has only a portion of \textit{comparison} type questions, whereas \strat{} comprises implied questions and comparison questions.

 Prior works have demonstrated that manual prompt designs are sensitive to change in order, sample size, and dataset \cite{zhao2021calibrate, lu-etal-2022-fantastically}. Hence, we propose an automated exemplar selection approach to select examples with similar token embedding distribution to the test sample.

\subsection{Dynamic Exemplar Selection}
\label{sec:dynamicselection}

The choice of exemplars for ICL plays a significant role in the downstream task performance. 
The interplay between each chosen demonstration and the test sample and the interplay between the demonstration samples affect the final performance. 
We hypothesize that it is important to select exemplars similar to the test sample but also ensure the chosen samples are not redundant concerning the test sample.
Hence, we require a selection metric that captures both quality and diversity when selecting an exemplar. Moreover, the selection metric should not be limited to capturing surface-level features (local consistency) but also semantic and global structures (global consistency). It should also not be sensitive to the choice of embedding model used to represent the test sample and the exemplars. 
We formalize this requirement on the selection metric $\gamma$ using the following \textbf{approximate invariance} property:



\begin{definition}
Let $E_\theta$ or $E_\phi$ be any two vector representation methods.
    Any distance metric $\gamma: \mathbb{R}^n \times \mathbb{R}^n \xrightarrow[]{}R$ is said to be approximately invariant, if 
    \begin{equation}
    \left|\left|\gamma\left(E_\phi\left(q_i^s\right),E_\phi\left(q_i^t\right)\right) - \gamma\left(E_\theta\left(q_i^s\right),E_\theta\left(q_i^t\right)\right)\right|\right| \leq \epsilon
\end{equation}
\end{definition}
In the next subsection, we propose a dynamic exemplar selection method satisfying the above property.

\subsubsection{Frechet term Embedding Distance (\ftd{}) for Exemplar Selection}

To select high-quality and diverse exemplars, we propose a new metric named Frechet Term Embedding Distance (FTD) inspired by Frechet Inception Distance \cite{heusel2018gans} and Frechet Embedding Distance \cite{frechet_language}. It has been shown that Frechet distance is relatively less sensitive to the choice of vector representation method \cite{semeniuta2019accurate,frechet_language} satisfying our invariance properties.
Unlike existing methods, our metric is at the per-instance level, measuring the diversity and quality of one exemplar when compared to the test example. Our approach is based on the intuition that the term embedding distributions of the exemplar from the transfer dataset should be similar to the distribution of term embeddings of the test example.






To simultaneously consider quality and diversity, we adapt the Frechet Inception Distance with a transformer-based model as the backbone. For each, $q_i^t$ we compute term-level vector representations using a transformer-based encoder model like $\mathbb{BERT}$ ($E_{\phi}$). Each example $q_i^s$ in $\psi_S$ is also projected to the vector space using the same encoder: $\left\{ e_i^t  ... e_{|q_i^t|}^t \right\}  = \mathtt{E_{\phi}}\left(q_i^t\right), \,\, \left\{ e_i^s(j) ...,e_{|q_i^s|}\right\}  = \mathtt{E_{\phi}}(q_i^s) \nonumber $
where $\{ e_i \}$ is a set of $d$-dimensional token embeddings from question $q_i^t$. Then the distributions over the representations are computed as,
$(\mu_i^t,\Sigma_i^t) = {NormEst}(\{ e_i^t  ... e_{|q_i^t|}^t \}) ,\,\, (\mu_i^s,\Sigma_i^s) = {NormEst}(\{ e_i^s  ... e_{|q_i^s|}^s) \}) 
$
, where NormEst estimates the mean and covariance matrix of a multivariate normal distribution fitted to the set of embeddings. Here, we assume that the term-level representations follow a multivariate Gaussian distribution. The distance between the token level distributions in $q_i^s$ and $q_i^t$ is computed using Frechet distance \cite{frechet}. We select the multivariate Gaussian distribution as it is invariant under transformations to underlying vectors. This is because the transformed vectors also follow a multivariate Gaussian distribution.
 If $X$ has a n-dimensional normal distribution with mean $\mu$ and $\Sigma$ covariance and suppose $v \in \mathbb{R}^m$ and $V \in\mathbb{R}^{m \times n} $, then $Y=v + VX$ is also a m-dimensional multivariate normal distribution.
 
The distance is computed  between the distribution $S$ with variables $(\mu_s,\Sigma_s)$ and $T$ with variables $(\mu_t,\Sigma_t)$ where $mu_t$ and $mu_s$ are of dimension (d)  obtained from token level representations of $q_i^s$ and $q_i^t$ respectively. 
We term it Frechet Term Distance (\ftd{}) computed as :
 \[\gamma = d_{FTD}\left(S,T \right) = \left|\left|\mu_i^t-\mu_i^s\right|\right| + Tr\left(\Sigma_i^s+\Sigma_i^t-2 \sqrt{\Sigma_i^s . \Sigma_i^t}\right)\]

 In the above expression, the first term computes the distance between the center of two distributions and the second term computes a metric over the space of covariance matrices. The above distance is computed at the sample level between all pairs formed between the test sample $q_i^t$ and all transfer dataset samples$\{q_1^s,q_2^s...\}$. Then the top exemplars with low distance to $q_i^t$ are selected in $\psi_S$ becoming prompts  for ICL. Since we compute mean and covariance across term embeddings, the mean represents the salient attribute across terms (global consistency) and the covariance characterizes interplay among terms. For instance, an exemplar that covers diverse topics would have high variance values.

 Since a multivariate normal distribution is invariant to affine transformations, such invariance also applies to variability in features \cite{invariant_features}. 
Hence, apart from $\mathbb{BERT}$ ($E_{\phi}$) even if a different encoder $E_{\theta}$ is used to produce contextualized embeddings, the resulting term distributions would also be a multivariate normal distribution.  Additionally, Frechet distance is equivalent to \textbf{Wasserstein distance} ($W$)\cite{frechet_language}. If a different encoder $E_{\theta}$ results in a slight shift in distribution parameters $\hat{S}$ and $\hat{T}$, then Wasserstein distance between the original and shifted distribution is bounded by $\epsilon$, $W(S,\hat{S}) \leq \epsilon$ and $W(T,\hat{T})\leq \epsilon$ \cite{kumar2023provable}. Then following the continuity bound proposed in \cite{arjovsky2017wasserstein} and extended in \cite{makkuva2020optimal} we derive, 
\begin{equation}
\label{eq:wass}
    \left|\left|W\left(S,T\right) - W\left(\hat{S},\hat{T}\right)\right|\right| \leq W\left(S,\hat{S}\right) \ \leq W\left(T,\hat{T}\right) \leq \epsilon
\end{equation} 
Equation \ref{eq:wass} directly implies $\left|\left|d_{\ftd{}}\left(S,T\right) - d_{\ftd{}}\left(\hat{S},\hat{T}\right)\right|\right| \leq \epsilon$.
Using this result and the fact that we compute mean and covariance across term embeddings, we can conclude that the proposed metric \ftd{} is less sensitive to choice of embedding, also supported by \cite{frechet_language}. Hence, this supports our notion of \textbf{approximate invariance} in Equation 2.
  We observe that FTD helps select good exemplars, as demonstrated by downstream task performance. 

%% file: table-datasets.tex
\begin{table*}[ht!]
\begin{tabular}{lllll}
\hline
\textbf{Dataset} & \textbf{Example Question} & \textbf{Description} & \textbf{Ability} & \textbf{Source}\\ \hline

\textbf{\arith{}}~\cite{roy2016solving} & \texttt{In fourth grade there were 10 students } & multi-step  & RG & \aqua{} \\
& \texttt{at the start of the year. During the year } & arithmetic word  & & \small{\textcolor{gray}{JS = 0.364}}\\
& \texttt{4 students left and 42 new students came to school.} &problems & &   \\
& How many students were in fourth grade at the end? & & &   \\

\textbf{\strat{}}~\cite{strategy_qa} \colorg & \texttt{Does Andrew Johnson's presidential number exceed}     \colorg & comparison and \colorg& QD \colorg& -- \colorg\\
\colorg& \texttt{Elagabalus's Emperor number?} \colorg &   compositional \colorg& \colorg&\colorg \\
\colorg& \texttt{} \colorg & \colorg questions& \colorg&\colorg \\
\textbf{\sv{}}~\cite{patel2021nlp} &  \texttt{Each pack of dvds costs 76 dollars.If there is a}      &  multi-step  &  RG & \aqua{} \\
&  \texttt{discount of 25 dollars on each pack. How much} & aithmetic word  & & \small{\textcolor{gray}{JS = 0.38}}\\
&  \texttt{do you have to pay to buy each pack?} & problems & & \\

\colorg \textbf{\sub{}}~\cite{tang-etal-2021-multi} & \colorg \texttt{What award did the writer of Never Let Me Go novel}      & \colorg compositional   & \colorg QD & \colorg -- \\
\colorg & \colorg \texttt{win in 1989?} & questions \colorg&\colorg & \colorg\\

\textbf{WQA}~\cite{2wikimultihopqa} & \texttt{Who was born later, Gideon Johnson or Holm Jølsen?}      & comparison and   & QD & \strat{} \\
 & & compositional&  & \sub{} \\
  & & questions &  & \small{\textcolor{gray}{JS = 0.33,0.57}} \\

\colorg \textbf{\aqua{}}~\cite{aqua_rat} & \colorg \texttt{A trader sold an article at a profit of 20\% for Rs.360.}      & \colorg  multi-step  & \colorg  RG & \colorg -- \\
\colorg  & \colorg  \texttt{What is the cost price of the article?} &\colorg arithmetic word & \colorg & \colorg \\
\colorg  & \colorg  \texttt{} &\colorg problems & \colorg & \colorg \\

 \textbf{\mqa{}}~\cite{trivedi-etal-2022-musique} &  \texttt{What did the actress in My Fair Lady win a Tony for ?}      &  compositional   &QD & \sub{}\\
& &questions & & \small{\textcolor{gray}{JS = 0.23}} \\

\colorg  \textbf{\tab{}}~\cite{tabmwp} & \colorg  \texttt{Allie kept track of how many kilometers she walked }      & \colorg Table based   &\colorg  RG &\colorg  -- \\
\colorg  & \colorg  \texttt{during the past 5 days. What is the range of the numbers?} &\colorg numerical reasoning   & \colorg & \colorg \\

\textbf{\fin{}}~\cite{chen2022finqa} & \texttt{In 2010 and 2009 , what was the total fair value }      & Table and Text   & RG & \tab{} \\
 & \texttt{in billions of assets segregated for the benefit }      & based   &  & \small{\textcolor{gray}{JS = 0.39}} \\
& \texttt{of securities and futures brokerage customers?}      & numerical reasoning&  &
\\

\bottomrule
\end{tabular}
\caption{Overview of the Complex QA or Abilities datasets used in this study. QD - Question Decomposition, RG - Rationale Generation for numerical reasoning.}
\label{tab:datasets_overview}
\end{table*}

%% file: 05-experiments-avi.tex
\section{Experimental Setup}
\label{sec:setup}

We reiterate that our aim is not to beat the state-of-the-art in each of the complex QA datasets.
Instead, our philosophy is to evaluate the degree to which we can use LLMs as general-purpose machines by (a) few-shot prompting, and (b) by re-using rationales from ability datasets.
Note that our variants do not involve any training, involving careful selection using pre-trained representations at best. 
With this in mind, in our experimental evaluation, we intend to answer the following research questions.


\mpara{RQ I.} Are the transferred abilities able to achieve competitive performance compared to 
manual prompting?

\mpara{RQ II.} Does automatic exemplar retrieval improve performance of \icat{} ?

\mpara{RQ III.} How does the choice of transfer datasets, and LLM substrates affect the performance of \icat{} ?

\subsection{Datasets, Setup, Metrics} 

To answer these research questions we conduct extensive experimentation over a wide variety of datasets covering diverse domains, abilities, and complexity types. 
The target datasets (and their dependent transfer datasets) are detailed in Table~\ref{tab:datasets_overview}. 
More details about the individual datasets used can be found in the supplementary material in the datasets section.

\subsubsection{LLM settings}
All our experiments are primarily carried out with instruction fine-tuned LLMs like \textit{gpt-3.5-turbo} unless otherwise specified. 
For all experiments, we set the temperature to $0.3$ to mitigate randomness, with frequency and presence penalty set to 0.8 and $0.6$ to avoid repetition. 
We cap the max\_token\_length to 900.  For all datasets, we use k=6 for all auto-exemplar selection methods.

\subsubsection{Metrics}
We use the official metrics of the target datasets, i.e., \textit{exact match}(EM) for \arith{}, \sv{}, \fin{}, and \textit{cover-EM} for~\wqa{} and~\mqa{}~\cite{cover_em, self_ask}.

\subsection{Baselines}
\label{sec:baselines}

Since our entire contribution is about using LLMs out-of-the-box for complex QA tasks, we start by two commonly acknowledged strong baselines -- Few-shot-COT, and Zero-shot-cot.

\textbf{Few-shot-COT}: We follow the hand annotated examples provided in prior works \cite{wei2023chainofthought, self_ask} as exemplars for ICL. 
For numerical reasoning this involves hand-annotated step-by-step reasoning chains and for complex QA the annotations are in the form of explanations to arrive at the final answer. 
The manual annotation limits the choice of examples and is also cumbersome.

\textbf{Zero-shot-COT}: The zero-shot chain-of-thought (COT) introduced by \cite{kojima2023large} follows a simple prompt prefix \textit{``Let's think step by step"} to induce reasoning abilities in LLMs. 

\textbf{Non-COT baseline}: We also use \fshot{} prompting~\cite{brown2020gpt3} as baseline, which does not involve writing explicit reasoning chains.

\subsubsection{Baselines optimized for Complex QA} Apart from the above general-purpose baselines, we consider two recent baselines that are optimized for complex QA using LLMs as substrates, i.e., Self-Ask~\cite{self_ask} and AutoCOT\cite{zhang2023automatic}. We provide details of these baselines in the baselines section in the supplementary.


\subsubsection{Auto Exemplar Selection} The final set of methods are ablations of our approach that experiment with different automated exemplar selection methods. In our setting, the exemplars are selected from a different distribution (\transfer{} datasets) unlike existing methods.

\textbf{KNN} \cite{rubin-etal-2022-learning}: We utilize the k-nearest neighbors-based function to select exemplars from the transfer dataset. The test sample and samples in transfer dataset are encoded using sentence transformer \textit{paraphrase-MiniLM-L6-v2}. 

\textbf{MMR} \cite{ye-etal-2023-complementary}: To induce diversity in the selected exemplars, the Maximum Marginal Relevance commonly used in IR \cite{carbonell1998use} is employed to select exemplar instances similar to $q_i^t$ but diverse with respect to exemplars already selected. We find $\lambda=7$ to be the optimal value for MMR based on evaluation on held out set.

\textbf{Random}: We also consider the scenario of random exemplar selection. We consider two cases, namely the static random and dynamic random, where the exemplars are selected once and for the latter case on a per-instance basis respectively.

%% file: 06-results-AA.tex
\section{Results}
\label{sec:results}
\input{table-results-v2}

From Table \ref{tab:main_result}, we observe that across tasks, the proposed \icat{} paradigm outperforms or achieves competitive performance with the Few-shot Chain of thought approach or other task-specific in-context learning methods. Our goal is not to beat the state-of-the-art, but to employ an efficient reasoning process without the overhead of manual annotations. We also plot the confusion matrices for the full evaluation sets for \icat{} vs \COT{}, as shown in the supplementary. Note, that in Table \ref{tab:main_result}, supervised SOTA results are included for completion and not for comparison, as it would not be a fair comparison to \icat{} which does not involve any training or supervision.

\subsection{Effectiveness of static selection}

In Table~\ref{tab:main_result}, we first observe that \icat{} with static selection or \icat{}-static outperforms the zero-shot-cot baseline. 
This is understandable for complex questions due to the lack of demonstrations.

However, we also find that performance of \icat{}-static is competitive or outperforms manual few-shot chain of thought or \COT{}~\cite{wei2023chainofthought}. 
This demonstrates that careful and manual selection of exemplars from transfer datasets that cover diverse categories of questions is \textit{already} sufficient for generating good decompositions. 
Noteworthy is that this diverse yet pre-selected static set of exemplars are competitive in most complex QA tasks.
Moreso, \icat{}-static even outperforms Auto-COT~\cite{zhang2023automatic}, a more recent approach to generate rationales for chosen exemplars by using zero-shot-COT. 
On manual inspection we found that the decline in performance is due to mistakes in rationales generated through zero-shot-COT. For instance, for a question

\mpara{Q:}\texttt{ 5 children were riding on the bus. At the bus stop, 63 children got off the bus. Then there were 14 children. How many more children got on the bus than those that got off?} 

the generated rationale is \textit{``Rationale: Let's think step by step. There were 63 children that got off the bus. That means there were 68 children on the bus before the 63 got off. 14 children remained on the bus, which means 54 children got on the bus. Therefore, 54 - 63 = -9. 9 more children got on the bus than those that got off. The answer is 9"}

Though the answer is correct the rationale is wrong as the operation should be ``14-9". 
This shows that the rationale is not necessarily faithful to the model's reasoning process in zero-shot-COT and may contain errors in reasoning. 
In qualitative analysis on \arith{} and \sv{}, we observe that \icat{} performs well on compositional reasoning, which requires performing multiple dependent arithmetic operations. We observe that in most of the erroneous instances in output of \COT{} are such multistep arithmetic reasoning questions. 
\input{table-ablations}
For complex QA, we observe that on \wqa{} we are able to achieve competitive performance only when transferring from both \sub{} and \strat{} and the performance drops when transferring only from either of the datasets in isolation as shown in Table \ref{tab:ablations}. 
We also observe that for \COT{}, at times the rationales generate factually inconsistent explanations, resulting in an incorrect answer, as shown in Table~\ref{tab:qualitative}. 
It has been observed that question decomposition is faithful to model reasoning process \cite{decomposing_complex_multi_hop} and also generates simpler sub-questions for reasoning. 

Examples from \wqa{} and \mqa{} are shown in Table \ref{tab:qualitative}. We observe that the first example hallucinates \textit{``Julia Andrews"} as the answer, whereas the decomposition by \icat{} results in the correct answer. Similarly, in the second example in the table we observe that \COT{} is unable to retrieve the birthdate of ``Holm Jolsen" from parametric memory, whereas our approach is able to correctly retrieve the answer.
\todo{Explanation/rationale fidelity seems to be out-of-context, we need to introduce it. Also merge this subsection with earlier sections -- we can have one para on questions decomp and another for numerical reasoning instead of having two separate subsections.}
We attribute the performance of \icat{}-static and \icat{} + \ftd{} to this observation, as the transfer datasets contain expert annotated sub-questions.

\input{table-qualitative}

\subsection{Effectiveness of dynamic selection}

Next, we evaluate auto-exemplar selection methods that aims to select exemplars from transfer datasets using a uncertainty-aware distance measure \ftd{}.
Note that, the exemplars selected from the \transfer{} dataset
are annotated with expert rationales which enable to select of few-shot examples with rationales unlike manual \COT{}, where randomly chosen samples are manually annotated with rationales limiting the pool to select samples from. 
We observe that the proposed uncertainty-aware term embedding distribution-based metric \ftd{} offers superior performance across \sv{}  and \arith{} and is competitive or offers gains over existing auto-exemplar selection approaches like $k$NN or MMR and is better than random selection approaches. 
Interestingly, \icat{}-static is also competitive or slightly underperforms compared to dynamic selection for \arith{} and \sv{}. 
Similar observations have been made in~\cite{zhou-etal-2023-revisiting} that careful manual selection of exemplars are sometimes competitive with dynamic selection. 
However, dynamic selection methods tend to more robust to perturbations in order, dataset and models \cite{zhao2021calibrate,su2022selective, lu-etal-2022-fantastically}. 
Overall, we observe that our best dynamic variant \icat{}-\ftd{} in general, selected representative and diverse samples that covered different categories of numerical reasoning problems which might have positively contributed to the rationale and solution for the test problem. 

For \wqa{} and \mqa{}, among dynamic selection approaches, we observe that the proposed \ftd{} based exemplar selection (\icat{} + (\ftd{})) outperforms static selection and random selection. 
It also outperforms exemplar selection  methods like KNN and MMR on \wqa{} dataset and offers slight gains on \mqa{} dataset. We observe that the results from \ftd{} are more consistent and stable compared to other dynamic selection methods or static selection methods across datasets.
We attribute this to the ability of FTD to select high-quality and diverse exemplars, ensuring global consistency \cite{frechet_language} from transfer datasets based on term distribution distance between the test sample and candidates in transfer datasets. 
Whereas, \knn{} and \mmr{} rely on local surface level characteristics of the instances. 
On examining the selected exemplars from \ftd{}, we observe that they matched the thematic style (location based, temporal questions etc..) of the test question and also the type of question (\textit{compositional, comparison}). Hence, we posit that representative, diverse exemplars with high quality annotation of reasoning steps like rationales or sub-questions induce faithful reasoning in LLMs for new questions. 
\ftd{} is also less sensitive to choice of embedding model \cite{frechet_language}, unlike \knn{} and \mmr{}.





\vspace{-1em}
\subsection{Impact of Task/Ability relatedness}
In this experiment, we evaluate the impact of the \textit{relatedness} of the transfer dataset to the target on the final task performance.
We operationalize this experiment on \fin{} that  requires reasoning over table and text, and we observe that the ability to parse table information to retrieve the appropriate numerical quantities are critical to the task.
We compare the \fin{} $\leftarrow$ \aqua{} against \fin{} $\leftarrow$ \tab{}.
It is clear from Table~\ref{tab:ablations} that transferring from \tab{} outperforms transferring from \aqua{}.
Unlike \aqua{}, \tab{} posseses the ability to parse tables along with rationale generation that is critical to the \fin{}.
An example of a \fin{} question is shown in Table \ref{tab:qualitative}. 
We observe that it is a \textit{compositional heterogeneous numerical reasoning question}. 
This requires parsing the correct information regarding securities and brokerages for $2009$ and $2010$ from table and text and adding them up to arrive at the final answer. We observe that the \COT{} does not take the securities information into account and misses performing addition of these values. However, \icat{} and \icat{} + \ftd{} are able to arrive at the correct answer due to the quality of rationales in the \transfer{} dataset \tab{}. 
We observe that the exemplars selected from \tab{} by \icat{} details the fields to access in the table for the required information to solve the problem, followed by the final algebraic expression to arrive at the solution. Since a significant portion of \fin{} (62.43\%) relies on information in tables for numerical reasoning, we posit transfer from \tab{} aids in increased performance.
Hence, we posit that in lack of similar tasks for transfer, datasets that contain desired abilities to solve parts of the task are still useful in a few-shot ICL setting. An example of prompt is shown in the prompts section in the supplementary.
\vspace{-1em}
\subsection{Impact of the different LLMs substrate}
\label{sec:llms}
We also evaluate the proposed approach \icat{} using open-source models like \flan{} (0.8B) and \flanxl{} (3B) to study the impact of scale and other properties of LLMs employed. We observe from Table \ref{tab:main_result} that scale of the LLM is critical apart from instruction based pre-training. This is due to the emergent capabilities being more pronounced in models of higher parameter scales \cite{wei2022emergent}. We also observe that models trained with high quality instructions perform better at diverse reasoning tasks. This is evident from \icat{} with gpt-3.5-turbo as backbone outperforming text-davinci-003 as shown in Table \ref{tab:main_result}.
\vspace{-1em}
\section{Conclusion}
In this work, we propose an approach \icat{} for using LLMs are general purpose machines to solve complex QA by transferring decomposition or rationales from diverse datasets. Our work obviates the need for manual annotation of chain of thought rationales adopted in existing methods like \COT{} and promotes reuse of existing datasets through transfer of ability, which is a characteristic of the datasets. We observe that the proposed \icat{} approach is competitive or outperforms existing few-shot prompting methods, and is adept at solving multistep or compositional questions. We also propose a new metric for automated exemplar selection approach \ftd{} to select demonstrations from \transfer{} datasets. We observe that the proposed metric is less sensitive to choice of embedding and provides robust performance across datasets, unlike existing approaches. In the future, we plan to explore more tasks in heterogeneous transfer like \fin{} when similar tasks are not available and also plan on grounding of answers from LLM.

\section{Ethical Considerations}

Since our approach uses LLMs for complex QA, the risks of hallucination \cite{hallucination} must be taken into consideration before deploying the approach in a QA system. Since users may trust the answers from the QA system hallucination, this may result in spread of misinformation \cite{zhang2023ethical,albrecht2022despite}. We observe that \icat{} results in less hallucinations compared to existing approaches due to step-by-step reasoning and transfer of expert rationales. Although hallucination is still a possibility when employing \icat{}. The grounding of answer generation on known knowledge sources by fact checking generated answers would help mitigate misinformation and dissemination of wrong facts through the QA system \cite{peng2023check}.

Additionally, we do not use any private information for the proposed approach. Though LLMs may have been pre-trained on sensitive information, our prompts do not elicit any sensitive information directly or indirectly. 

%% file: table-results-v2.tex
\begin{table}[!t]
    \centering
    
    \begin{tabular}{lccccc}
    \toprule
     \textbf{Method}& \multicolumn{1}{c}{MQA}& \multicolumn{1}{c}{\wqa{}} & \multicolumn{1}{c}{MArith} & \multicolumn{1}{c}{\sv{}}& \multicolumn{1}{c}{\fin{}}\\
     
     \midrule
     \fshot{}& 11.66& 28.41& 85.74& 78.2& 42.55\\
     
     \COT{} &20.85& 34.08&94.79&  79.4&52.22 \\
     
     \zfshot{} & 12.22& 19.25& 88.75& 75.4& 47.51 \\
    \hline
    \self{} & 23.64& 32.58& N/A& N/A& N/A \\
      
      \autocot{} &  N/A& N/A&  93.83 & 76.2 & N/A \\
     \midrule
\colorg \textbf{\icat{} (static)} & \colorg& \colorg& \colorg& \colorg& \colorg\\
     text-davinci &  20.28& 25.33&  81.71 & 74.7 & 40.01 \\
    
    \flan{} &  1.59 & 20.00 &  10.25 & 8.31 & 1.48 \\
      
          random &  23.24& 25.08&  92.44 & 75.90 & 51.17 \\
      
      \flanxl{} &  2.96 & 17.92 &  24.48 & 22.08 & 1.92 \\
      
      gpt-3.5-turbo &  25.24$\dagger$&  34.00& \colorit  96.14 $\dagger$ & \coloritt  80.0 & 53.35 \\

      \midrule
\colorg \textbf{\icat{} (dynamic)} &\colorg & \colorg& \colorg& \colorg& \colorg\\

   random&  24.36& 29.41&  92.95 &   77.60 & 50.74 \\

    \knn{}  &  26.99& 30.69&  \coloritt  95.13 & \colorit  80.70 & 52.65 \\
     \mmr{}  &  27.07& 30.33&  94.96 & \colorit  80.70 & 52.14 \\
    \ftd{} &  \coloritt  \textbf{27.24}$\ddagger$& \coloritt  \textbf{36.00}&  \colorit  \textbf{96.14}$\dagger$ & \colorit  \textbf{80.70} & \coloritt \textbf{55.36}$\dagger$ \\
    \midrule
    
    supervised SOTA & \bf \colorit 37.60  & \colorit  \bf 50.59 &   60.5  &  47.3 & \bf \colorit  58.86\\
     
     \bottomrule
    \end{tabular}
    \caption{Results across datasets. The model used for \icat{} and other few-shot approaches is gpt-3.5-turbo unless otherwise specified.$\dagger$ and $\ddagger$ indicates statistical significance over \COT{} at 0.1 level  and 0.01 level respectively. SOTA supervised models included for completion.}
    \vspace{-1cm}
    \label{tab:main_result}
\end{table}

%% file: table-ablations.tex
\begin{table}

    \begin{tabular}{lccccc}
    \toprule
     \textbf{\transfer{}}& \multicolumn{1}{c}{\textbf{\target{}}} & \textbf{EM}\\ \midrule
     \tab{} &   & \textbf{ 53.35 } \\
    \aqua{} & \fin{}  &  49.60  \\
     \midrule
    \sub{} &  &  31.08   \\

        \strat{} & \wqa{} &  30.58   \\
        
    \sub{} + \strat{} &  &  \textbf{34.00}  \\
     
     \bottomrule
    \end{tabular}
    
    \caption{Dataset ablations for \icat{} (static)}
    \vspace{-1cm}
    \label{tab:ablations}
\end{table}

%% file: table-qualitative.tex
\begin{table*}
\begin{tabular}{lp{.80\textwidth}}
\toprule
    \textbf{Method}     & \textbf{Decomposition} \\
\midrule
 \colorg \textbf{Question}  & \colorg \textbf{What did the actress in My Fair Lady win a Tony for? [Dataset: \textbf{MQA}]} \\

\COT{}  & The actress in My Fair Lady who won a Tony award is \textcolor{red}{Julie Andrews}. She won the Tony for Best Actress [\dots]  \\

\colorg \icat{} + \ftd{} & \colorg \textbf{[Question 0]}: Who is the actress in My Fair Lady?
\textbf{[Answer 0]:} \textcolor{teal}{\texttt{Audrey Hepburn}}
\textbf{[Question 1]}: What did Audrey Hepburn win a Tony for?
\textbf{[Answer 1]:} \texttt{Best Actress in a Musical Ondine}. \\
\icat{} & \textbf{[Question1]:} Who is the actress in My Fair Lady?
\textbf{[Answer1]}: \texttt{Audrey Hepburn}
\textbf{[Question2]:} What did Audrey Hepburn win a Tony for?
\textbf{[Answer2]}: \texttt{Audrey Hepburn did not win a Tony award.} \\
\midrule
\midrule
\colorg \textbf{Question}  & \textbf{Who was born later, Gideon Johnson Pillow or Holm Jølsen?} [Dataset: \textbf{MQA}] \colorg \\
\COT{}  & Gideon Johnson Pillow was born on June 8, 1806. \textcolor{red}{Holm Jølsen's birthdate is not provided.}  \\
\colorg \icat{} + \ftd{} & \colorg \textbf{[Question 0]}: When was Gideon Johnson Pillow born?
\textbf{[Answer 0]}: June 8, 1806
\textbf{[Question 1]}: When was Holm Jølsen born?
\textbf{[Answer 1]}: \textcolor{teal}{\texttt{October 5, 1852}}
\textbf{[Question 2]}: Who was born later, Gideon Johnson Pillow or Holm Jølsen?
\textbf{[Answer 2]}: \textcolor{teal}{\texttt{Holm Jølsen}} \\
\icat{} & \textbf{[Question 0]}: When was Gideon Johnson Pillow born?
\textbf{[Answer 0]}: \texttt{June 8, 1806}
\textbf{[Question 1]}: When was Holm Jølsen born?
\textbf{[Answer 1]}: \texttt{November 17, 1867}
\textbf{[Question 2]}: Who was born later, Gideon Johnson Pillow or Holm Jølsen?
\textbf{[Answer 2]}: Holm Jølsen \\
\midrule
\midrule

\colorg \textbf{Question}  & \colorg \textbf{In 2010 and 2009 , what was the total fair value in billions of assets segregated for the benefit of securities and futures brokerage customers?} [Dataset: \textbf{FinQA}] \\
\COT{}  & To find the total fair value in billions of assets segregated for the benefit of securities and futures brokerage customers in 2010 and 2009, we need to look at the table. \textcolor{red}{In 2010, the fair value was \$25.0 billion and in 2009, it was \$24.0 billion. Hence total is \textbf{Answer:} \$49 billion.}  \\
\colorg \icat{} + \ftd{} & \colorg \textbf{Rationale:} To find the total fair value in billions [\dots], look at the table for the years 2010 and 2009. In the table, under "year ended December 31," we can see two columns [\dots], there is a row labeled "securities and futures brokerage customers." In this row, there are values listed in millions.To convert these values to billions, we divide each value by 1000.
\textcolor{teal}{For 2010: \$25.0 billion (segregated cash) + \$9.7 billion (fair value of securities) = \$34.7 billion
For 2009: \$24.0 billion (segregated cash) + \$10.2 billion (fair value of securities) = \$34.2 billion
[\dots] \textbf{Answer:} 68.9}
 \\
\icat{} & in billions, we need to look at Note 29 in the text. 
In 2010, \$25.0 billion in cash and \$9.7 billion in securities with [\dots]. 
In 2009, \$24.0 billion in cash and \$10.2 billion in securities [\dots].
\textbf{Answer}: \texttt{68.9 (billion)}. \\
\midrule
\midrule

\end{tabular}
\caption{Qualitative analysis of reasoning chains generated by chain-of-thought vs \icat{}}
\label{tab:qualitative}
\end{table*}